\documentclass[pdflatex,sn-mathphys-num]{sn-jnl}

\usepackage[table]{xcolor}
\usepackage{graphicx}%
\usepackage{multirow}%
\usepackage{amsmath,amssymb,amsfonts}%
\usepackage{amsthm}%
\usepackage{mathrsfs}%
\usepackage[title]{appendix}%
\usepackage{textcomp}%
\usepackage{manyfoot}%
\usepackage{algorithm}%
\usepackage{algorithmicx}%
\usepackage{algpseudocode}%
\usepackage{listings}%
\usepackage{tabularx}
\usepackage[utf8]{inputenc} 
\usepackage[T1]{fontenc}    
\usepackage{hyperref}       
\usepackage{url}            
\usepackage{booktabs}       
\usepackage{nicefrac}       
\usepackage{microtype}      

\usepackage{listings}
\usepackage{twemojis}
\usepackage{graphicx}
\usepackage{xspace}
\usepackage{amsmath}
\usepackage{amssymb}

\usepackage{makecell} 
\usepackage{subfigure}
\usepackage{longtable}

\usepackage{wrapfig} 
\usepackage{float}
\usepackage{enumitem}
\usepackage{threeparttable}
\usepackage{forest}

\graphicspath{{./figures/}}

\definecolor{medblue}{RGB}{0,102,204}
\definecolor{lightblue}{cmyk}{0.1,0.0,0.02,0.02}
\newcommand{\bigemoji}[1]{\raisebox{0.05em}{\scalebox{2}{\twemoji{#1}}}}

\newcommand{\administrationandworkflowrange}{(0.53-0.63)}
\newcommand{\clinicaldecisionsupportrange}{(0.61-0.76)}
\newcommand{\clinicalnotegenerationrange}{(0.74-0.85)}
\newcommand{\medicalresearchassistancerange}{(0.65-0.75)}
\newcommand{\patientcommunicationandeducationrange}{(0.76-0.89)}

\newcommand{\framework}{\textbf{{MedHELM}}}

\theoremstyle{thmstyleone}%
\theoremstyle{thmstyletwo}%

\theoremstyle{thmstylethree}%

\begin{document}

\title{ \bigemoji{hospital} \framework: Holistic Evaluation of Large Language Models for Medical Tasks }


\author[1]{\fnm{Suhana}          \sur{Bedi}}
\equalcont{These authors contributed equally to this work and are listed in alphabetical order.}
\author[1]{\fnm{Hejie}           \sur{Cui}}
\equalcont{These authors contributed equally to this work and are listed in alphabetical order.}
\author[1]{\fnm{Miguel}    \sur{Fuentes}}
\equalcont{These authors contributed equally to this work and are listed in alphabetical order.}
\author[1]{\fnm{Alyssa}          \sur{Unell}}
\equalcont{These authors contributed equally to this work and are listed in alphabetical order.}
\author[1]{\fnm{Michael}         \sur{Wornow}}
\author[2]{\fnm{Juan M.}         \sur{Banda}}
\author[2]{\fnm{Nikesh}          \sur{Kotecha}}
\author[2]{\fnm{Timothy}         \sur{Keyes}}
\author[3]{\fnm{Yifan}           \sur{Mai}}
\author[4]{\fnm{Mert}            \sur{Oez}}
\author[4]{\fnm{Hao}             \sur{Qiu}}
\author[4]{\fnm{Shrey}           \sur{Jain}}
\author[4]{\fnm{Leonardo}        \sur{Schettini}}
\author[1]{\fnm{Mehr}           \sur{Kashyap}}
\author[1]{\fnm{Jason Alan}      \sur{Fries}}
\author[1]{\fnm{Akshay}          \sur{Swaminathan}}
\author[1]{\fnm{Philip}          \sur{Chung}}
\author[1]{\fnm{Fateme}          \sur{Nateghi}}
\author[1]{\fnm{Asad}            \sur{Aali}}
\author[1]{\fnm{Ashwin}          \sur{Nayak}}
\author[1]{\fnm{Shivam}          \sur{Vedak}}
\author[1]{\fnm{Sneha S.}           \sur{Jain}}
\author[1]{\fnm{Birju}           \sur{Patel}}
\author[1]{\fnm{Oluseyi}         \sur{Fayanju}}
\author[1]{\fnm{Shreya}       \sur{Shah}}
\author[1]{\fnm{Ethan}           \sur{Goh}}
\author[1]{\fnm{Dong-han}        \sur{Yao}}
\author[1]{\fnm{Brian}           \sur{Soetikno}}
\author[1]{\fnm{Eduardo}         \sur{Reis}}
\author[1]{\fnm{Sergios}         \sur{Gatidis}}
\author[1]{\fnm{Vasu}            \sur{Divi}}
\author[1]{\fnm{Robson}          \sur{Capasso}}
\author[1]{\fnm{Rachna}          \sur{Saralkar}}
\author[1]{\fnm{Chia-Chun}       \sur{Chiang}}
\author[1]{\fnm{Jenelle}         \sur{Jindal}}
\author[1]{\fnm{Tho}             \sur{Pham}}
\author[1]{\fnm{Faraz}           \sur{Ghoddusi}}
\author[1]{\fnm{Steven}          \sur{Lin}}
\author[1]{\fnm{Albert}          \sur{S. Chiou}}
\author[1]{\fnm{Christy}         \sur{Hong}}
\author[1]{\fnm{Mohana}          \sur{Roy}}
\author[1]{\fnm{Michael}         \sur{F. Gensheimer}}
\author[1]{\fnm{Hinesh}          \sur{Patel}}
\author[1]{\fnm{Kevin}           \sur{Schulman}}
\author[1]{\fnm{Dev}             \sur{Dash}}
\author[1]{\fnm{Danton}          \sur{Char}}
\author[1]{\fnm{Lance}           \sur{Downing}}
\author[1]{\fnm{Francois}        \sur{Grolleau}}
\author[1]{\fnm{Kameron}         \sur{Black}}
\author[1]{\fnm{Bethel}          \sur{Mieso}}
\author[1]{\fnm{Aydin}           \sur{Zahedivash}}
\author[4]{\fnm{Wen-wai}         \sur{Yim}}
\author[4]{\fnm{Harshita}        \sur{Sharma}}
\author[3]{\fnm{Tony}            \sur{Lee}}
\author[2]{\fnm{Hannah}          \sur{Kirsch}}
\author[2]{\fnm{Jennifer}        \sur{Lee}}
\author[2]{\fnm{Nerissa}         \sur{Ambers}}
\author[2]{\fnm{Carlene}         \sur{Lugtu}}
\author[2]{\fnm{Aditya}          \sur{Sharma}}
\author[2]{\fnm{Bilal}           \sur{Mawji}}
\author[2]{\fnm{Alex}            \sur{Alekseyev}}
\author[2]{\fnm{Vicky}           \sur{Zhou}}
\author[2]{\fnm{Vikas}           \sur{Kakkar}}
\author[2]{\fnm{Jarrod}          \sur{Helzer}}
\author[2]{\fnm{Anurang}         \sur{Revri}}
\author[1]{\fnm{Yair}            \sur{Bannett}}
\author[1]{\fnm{Roxana}          \sur{Daneshjou}}
\author[1]{\fnm{Jonathan}        \sur{Chen}}
\author[1]{\fnm{Emily}           \sur{Alsentzer}}
\author[1]{\fnm{Keith}           \sur{Morse}}
\author[1]{\fnm{Nirmal}          \sur{Ravi}}
\author[1]{\fnm{Nima}            \sur{Aghaeepour}}
\author[1]{\fnm{Vanessa}         \sur{Kennedy}}
\author[1]{\fnm{Akshay}          \sur{Chaudhari}}
\author[1,2]{\fnm{Thomas}           \sur{Wang}}
\author[3, 5]{\fnm{Sanmi}           \sur{Koyejo}}
\author[1,4]{\fnm{Matthew}         \sur{P. Lungren}}
\author[4, 5]{\fnm{Eric}            \sur{Horvitz}}
\author[3]{\fnm{Percy}           \sur{Liang}}
\author[2]{\fnm{Mike}            \sur{Pfeffer}}
\author*[1,2]{\fnm{Nigam H.}           \sur{Shah}}\email{nigam@stanford.edu}
\affil[1]{Stanford University School of Medicine, Stanford, CA, USA}
\affil[2]{Stanford Health Care, Palo Alto, CA, USA}
\affil[3]{Center for Research on Foundation Models (CRFM) \& Department of Computer Science, Stanford University, CA, USA}
\affil[4]{Microsoft Corporation, Redmond, WA, USA}
\affil[5]{Stanford Institute for Human-Centered AI, CA, USA}

\abstract{
While large language models (LLMs) achieve near-perfect scores on medical licensing exams, these evaluations
inadequately reflect the complexity and diversity of real-world clinical practice. We introduce MedHELM, an extensible evaluation framework for assessing LLM performance for medical tasks with three key contributions. First, we present a clinician-validated taxonomy spanning five categories, 22 subcategories, and 121 tasks developed with 29 clinicians. Second, we develop a comprehensive benchmark suite comprising 35 benchmarks (17 existing, 18 newly formulated) providing complete coverage of all categories and subcategories in the taxonomy. Third, we conduct a systematic comparison of LLMs with improved evaluation methods (using an LLM-jury) and a cost-performance analysis. Evaluation of nine frontier LLMs, using the 35 benchmarks, revealed significant performance variation. Advanced reasoning models (DeepSeek R1: 66\% win-rate; o3-mini: 64\% win-rate) demonstrated superior performance, though Claude 3.5 Sonnet achieved comparable results at 40\% lower estimated computational cost. On a normalized accuracy scale (0-1), most models performed strongly in Clinical Note Generation \clinicalnotegenerationrange\  and Patient Communication \& Education \patientcommunicationandeducationrange, moderately in Medical Research Assistance \medicalresearchassistancerange\ and Clinical Decision Support \clinicaldecisionsupportrange, and lower in Administration \& Workflow \administrationandworkflowrange. Our LLM-jury evaluation method achieved good agreement with clinician ratings (ICC = 0.47), surpassing both average clinician–clinician agreement (ICC = 0.43) and automated baselines, including ROUGE-L (0.36) and BERTScore-F1 (0.44). Claude 3.5 Sonnet achieved comparable performance to top models at lower estimated cost. These findings highlight the importance of real-world, task-specific evaluation for medical use of LLMs and provides an open source framework to enable this.
}




\maketitle

\section{Introduction}\label{sec1}

Large Language Models (LLMs) have shown impressive performance on medical knowledge benchmarks, achieving $\sim$99\% accuracy on standardized exams like MedQA \cite{paperswithcode_medqa}. This has sparked interest in deploying them in healthcare settings: supporting clinical decision-making such as diagnosis and treatment \cite{khosravi2024ai}, optimizing clinical workflows including documentation and scheduling \cite{nath2024ai}, and enhancing patient education and communication \cite{carl2025evaluating}. 

However, there is a large gap between performance on medical knowledge benchmarks and readiness for real-world deployment due to three key limitations in these existing benchmarks \cite{nori2023capabilitiesgpt4medicalchallenge}: 
(1) \textit{Questions do not match real-world settings} -- Existing benchmarks rely on synthetic vignettes or narrowly-scoped exam questions, failing to capture key aspects of real diagnostic processes such as extracting relevant details from patient records~\cite{raji2025bench, medmcqa}. (2) \textit{Limited use of real-world data} -- Only 5\% of LLM evaluations use real-world electronic health record (EHR) data \cite{bedi2025testing}. EHRs contain ambiguities, inconsistencies, and domain-specific shorthand that synthetic data cannot replicate. (3) \textit{Limited task diversity} -- Around 64\% of LLM evaluations in healthcare focus only on medical licensing exams and diagnostic tasks \cite{bedi2025testing}, ignoring essential hospital operations such as administrative tasks (e.g., generating prior authorization letters, identifying billing codes), clinical documentation (e.g., writing progress notes or discharge instructions), and patient communication (e.g., asynchronous messaging through electronic patient portals) \cite{hager2024evaluation}.

Recent work on HealthBench \citep{arora2025healthbench} has advanced the evaluation of LLMs in medicine by scoring 5000 single-turn, free-text dialogues in which the model acts independently, much like a direct-to-patient advice line, without follow-up questions or human oversight. Its physician-authored rubrics reward exhaustive, risk-averse responses that maximize safety and completeness, offering a valuable stress test for fully autonomous chatbots. However, this design does not capture the iterative, context-aware interactions clinicians expect from an assistive co-pilot, nor does it assess performance on structured tasks that dominate everyday workflows (e.g., order review, note generation, literature summarization).

\begin{figure}[t]
\centering
  \includegraphics[width=\textwidth]{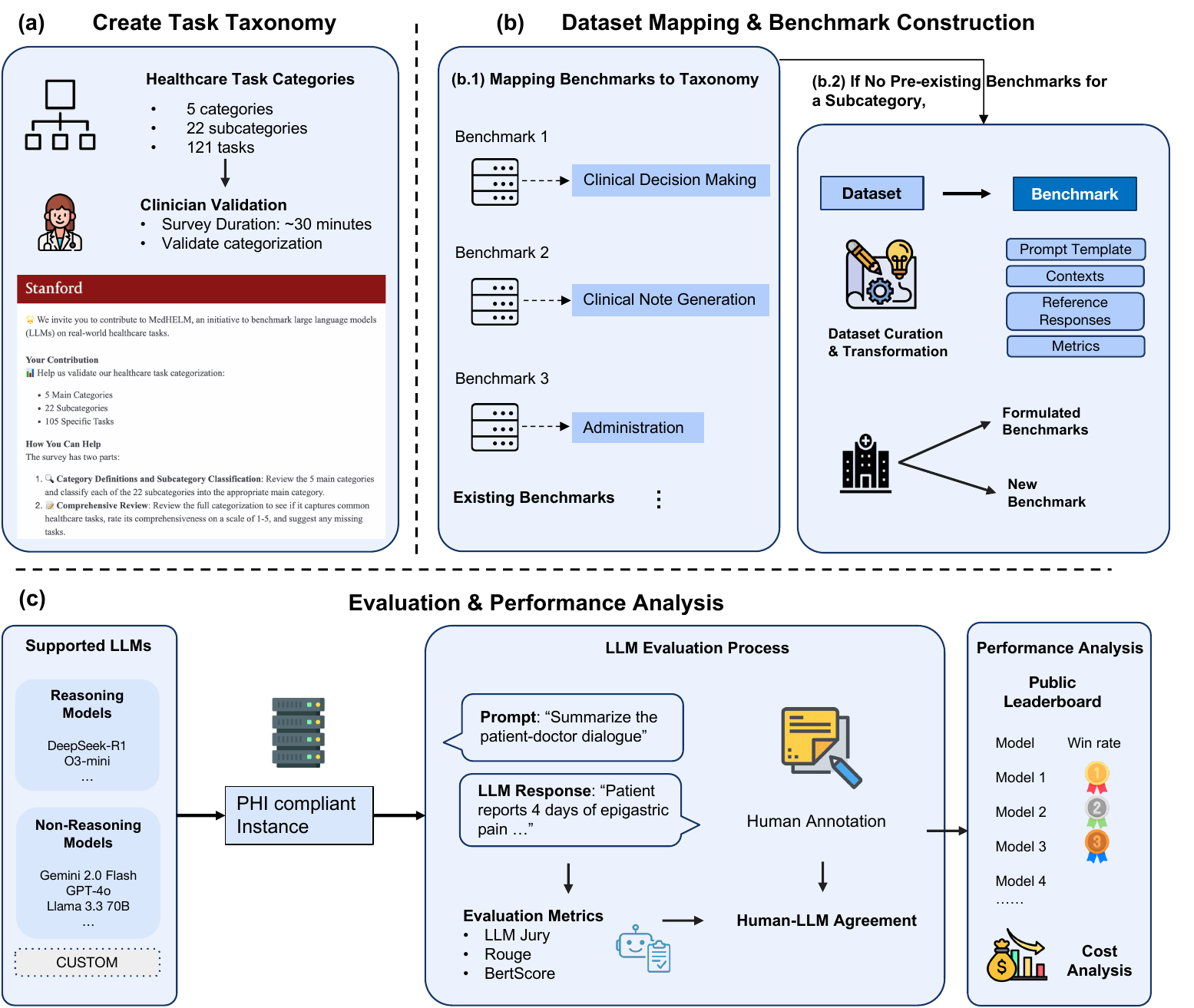}
    \caption{This figure illustrates: (a) a clinician-validated taxonomy organizing 121 medical tasks into five categories and 22 subcategories; (b) a suite of benchmarks that map existing benchmarks to this taxonomy and introduces new benchmarks for complete coverage; and (c) an evaluation comparing reasoning and non-reasoning LLMs, with model rankings, LLM jury based evaluation of open-ended benchmarks, and cost-performance analysis}
    \label{fig:framework}
\end{figure}

To address these limitations, we introduce \framework\ (Holistic Evaluation of Large Language Models for Medical Tasks), an extensible evaluation framework for assessing LLM performance in completing medical tasks (Figure \ref{fig:framework}). Inspired by the HELM project's standardized cross-domain evaluations \cite{liang2023holistic}, using \framework\, we evaluate nine LLMs using 35 distinct benchmarks covering all 22 subcategories of medical tasks, focusing on clinicians' day-to-day activities beyond just taking licensing exams. We assess performance using benchmark-appropriate metrics (i.e., exact match for closed-ended benchmarks, LLM-jury where three LLMs evaluate responses using tailored rubrics for open-ended benchmarks with demonstrated agreement to clinician ratings) as well as estimated computational cost to provide practical deployment insights. Our primary contributions are:

\vspace{0.5em}
\begin{enumerate}[label=\arabic*.,nosep,leftmargin=*]
    \item \textbf{Clinician‑validated taxonomy:} A five-category, 22-subcategory, 121-task taxonomy developed with 29 clinicians. Clinicians achieved a 96.7\% agreement rate when mapping subcategories to appropriate categories, validating the clear and discrete nature of the taxonomy. We present the complete taxonomy in the Results and Appendix sections.
    \item \textbf{A benchmark suite with full taxonomy coverage:} 
    A collection of 35 benchmarks spanning all 22 subcategories of medical tasks. This includes 17 existing benchmarks, five re-formulated benchmarks based on existing datasets, and 13 new benchmarks.\footnote{For privacy and regulatory compliance, as well as to prevent inclusion in LLM training data, 14 datasets are not publicly released.}
    \item \textbf{Comparative evaluation of models along with cost-performance analysis:} A systematic evaluation shows that reasoning models  achieve the highest overall performance. 
\end{enumerate}

The \framework\ framework addresses a critical need in the testing and evaluation of AI for medical use by providing consistent, real-world evaluation standards for medical (and healthcare) applications of LLMs. This framework benefits three key stakeholder groups: (1) Healthcare systems evaluating LLMs for specific tasks, (2) AI developers identifying performance gaps across medical tasks, and (3) Researchers developing methods to reproducibly measure LLM capabilities on medical tasks. To foster collaborative improvement of such AI evaluation, we provide an openly accessible 
\textbf{leaderboard}
\footnote{\url{https://crfm.stanford.edu/helm/medhelm/v2.0.0/ }} with current benchmarking results and share the \textbf{codebase} \footnote{\url{https://github.com/stanford-crfm/helm}}, with \textbf{documentation} \footnote{\url{https://crfm-helm.readthedocs.io/en/latest/medhelm/}} for contributing new datasets, evaluation metrics, and benchmarking custom models. By standardizing terminology and evaluation methods across the task taxonomy, \framework\ establishes a foundation for reproducible and real-world assessment of AI capabilities in medicine.

\section{Results}\label{sec2}

\subsection{Clinician Validation of the Taxonomy}
In a structured review process, 29 clinicians evaluated the initial taxonomy comprising five categories, 21 subcategories, and 98 tasks. When asked to assign each subcategory to its appropriate top-level category, clinicians correctly matched 96.7\% of subcategories to their intended categories. The clinicians rated the comprehensiveness of the proposed tasks at a mean of 4.21/5 (n = 29) and provided 107 comments with suggestions for improvement. Based on this feedback, we refined task definitions and expanded the taxonomy to its final form: five categories, 22 subcategories, and 121 tasks.  An overview of the final taxonomy is shown in Figure \ref{fig:taxonomy}, with the complete list provided in Appendix~\ref{app:taxonomy}. 

\begin{figure*}[t]
  \centering
  \includegraphics[width=0.95\linewidth]{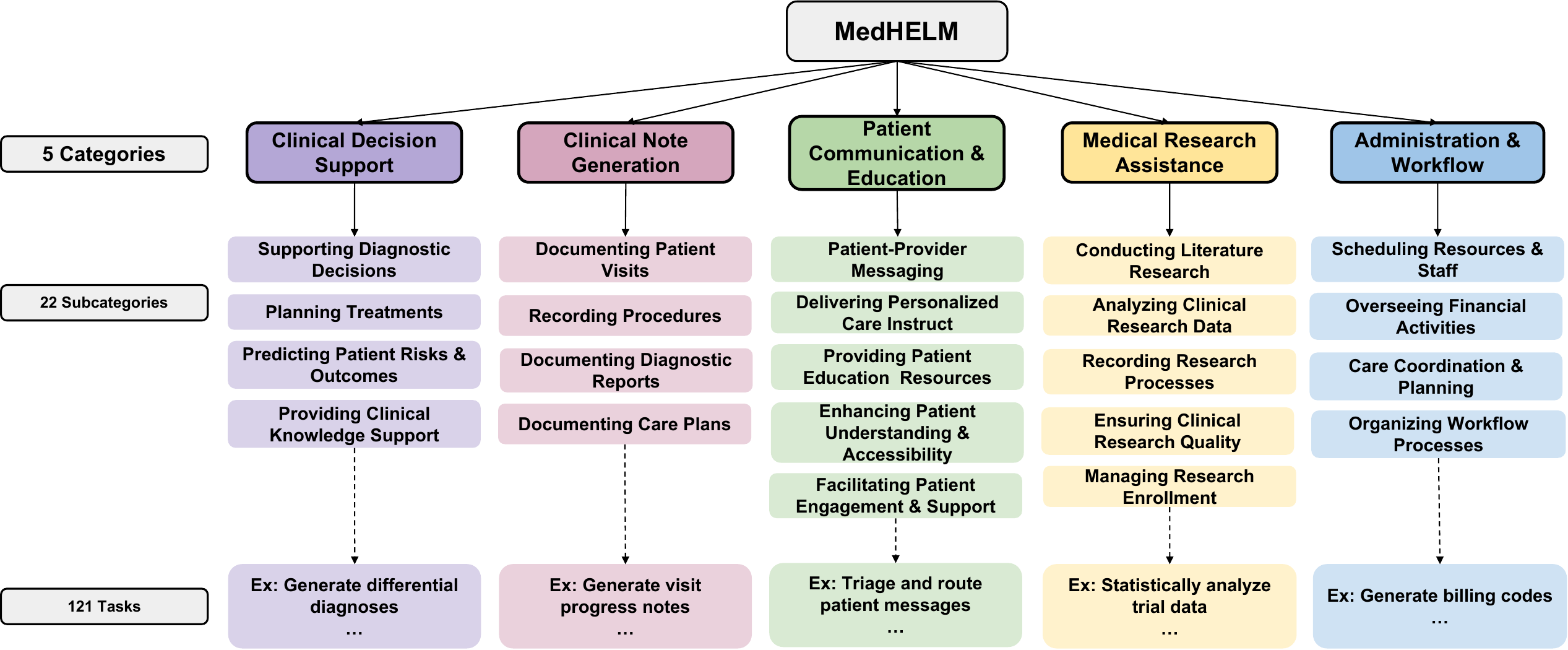}
  \caption{Overview of the final taxonomy comprising five main categories and 22 subcategories.}
  \label{fig:taxonomy}
  \vskip -0.8em
\end{figure*}

\subsection{Overview of the Benchmark Suite}
\label{ssec:benchmark_overview}

Our 35 benchmarks span all 22 subcategories, providing full coverage over categories and subcategories in our taxonomy (Table \ref{tab:new_datasets}). The benchmark suite comprises 17 \textit{existing} benchmarks, five \textit{re-formulated} benchmarks derived from previously unevaluated medical datasets, and 13 \textit{new} benchmarks, out of which 12 are EHR-based. The suite includes 13 open-ended benchmarks (requiring free-text generation) and 22 closed-ended benchmarks (with predefined answer choices). Access levels are designated as 14 public, seven gated (i.e. requiring approval), and 14 private.

\textit{Clinical Decision Support} is the most represented category with ten benchmarks, followed by \textit{Patient Communication} (eight), \textit{Clinical Note Generation} and \textit{Medical Research Assistance} (six), and \textit{Administration \& Workflow} (five). The distribution of benchmarks across subcategories is uneven, with 15 subcategories containing a single benchmark and the remaining seven subcategories containing between two and five benchmarks each.

\subsection{Model Evaluation \& Cost-Performance Analysis}
\label{ssec:q1}

\subsubsection{Overall Performance}
\paragraph{Pairwise Win-Rate and Average Scores}
Table~\ref{tab:model_macroavg} compares models using win-rate and macro-average performance metrics (defined in the table caption). DeepSeek R1 performed best, winning 66\% of head-to-head comparisons with a macro-average of 0.75 and low win standard deviation (0.10). o3-mini followed with a 64\% win rate and the highest macro-average (0.77), driven by strong performance in benchmarks in the clinical decision support category. 

The Claude models achieved 63-64\% win rates and identical macro-averages (0.73). GPT-4o achieved a 57\% win rate, while Gemini 2.0 Flash (42\%) and GPT-4o mini (39\%) performed lower. Open-source Llama 3.3 Instruct achieved a 30\% win rate. Gemini 1.5 Pro ranked lowest with 24\% wins but had the lowest win standard deviation (0.08), showing the most consistent competitive performance.

\begin{table}[ht]
  \centering
  \small
  \begin{tabular}{lccccc}
\toprule
Model (snapshot) & Win-rate$\uparrow$ & Win SD$\downarrow$ & Macro-avg$\uparrow$ & SD$\downarrow$ \\
\midrule
DeepSeek R1 & \textbf{0.66} & 0.10 & 0.75 & 0.22 \\
o3-mini (2025-01-31) & 0.64 & 0.16 & \textbf{0.77} & \textbf{0.18} \\
Claude 3.7 Sonnet (20250219) & 0.64 & 0.13 & 0.73 & 0.21 \\
Claude 3.5 Sonnet (20241022) & 0.63 & 0.14 & 0.73 & 0.21 \\
GPT-4o (2024-05-13) & 0.57 & 0.17 & 0.73 & 0.18 \\
Gemini 2.0 Flash & 0.42 & 0.17 & 0.70 & 0.21 \\
GPT-4o mini (2024-07-18) & 0.39 & 0.18 & 0.71 & 0.20 \\
Llama 3.3 Instruct (70B) & 0.30 & 0.13 & 0.69 & 0.22 \\
Gemini 1.5 Pro (001) & 0.24 & \textbf{0.08} & 0.67 & 0.21 \\
\bottomrule
\end{tabular}
  \caption{%
    Comparison of performance of frontier models across 35 MedHELM benchmarks,
    sorted by descending win‑rate. \textbf{Bold} indicates the best value in
    each column. Win-rate represents the proportion of pairwise
comparisons where each model achieved superior performance across all
35 benchmarks (possible range: 0-1). Win standard deviation (SD) measures how consistently a model wins (lower values = more consistent). Macro-avg is the average performance score across all 35 benchmarks. SD shows how much performance varies across different benchmarks (lower values = more consistent across benchmarks). 
    }
  \label{tab:model_macroavg}

  \vskip -1.0em
\end{table}

\begin{figure}[t]
\centering
  \includegraphics[width=0.9\textwidth]{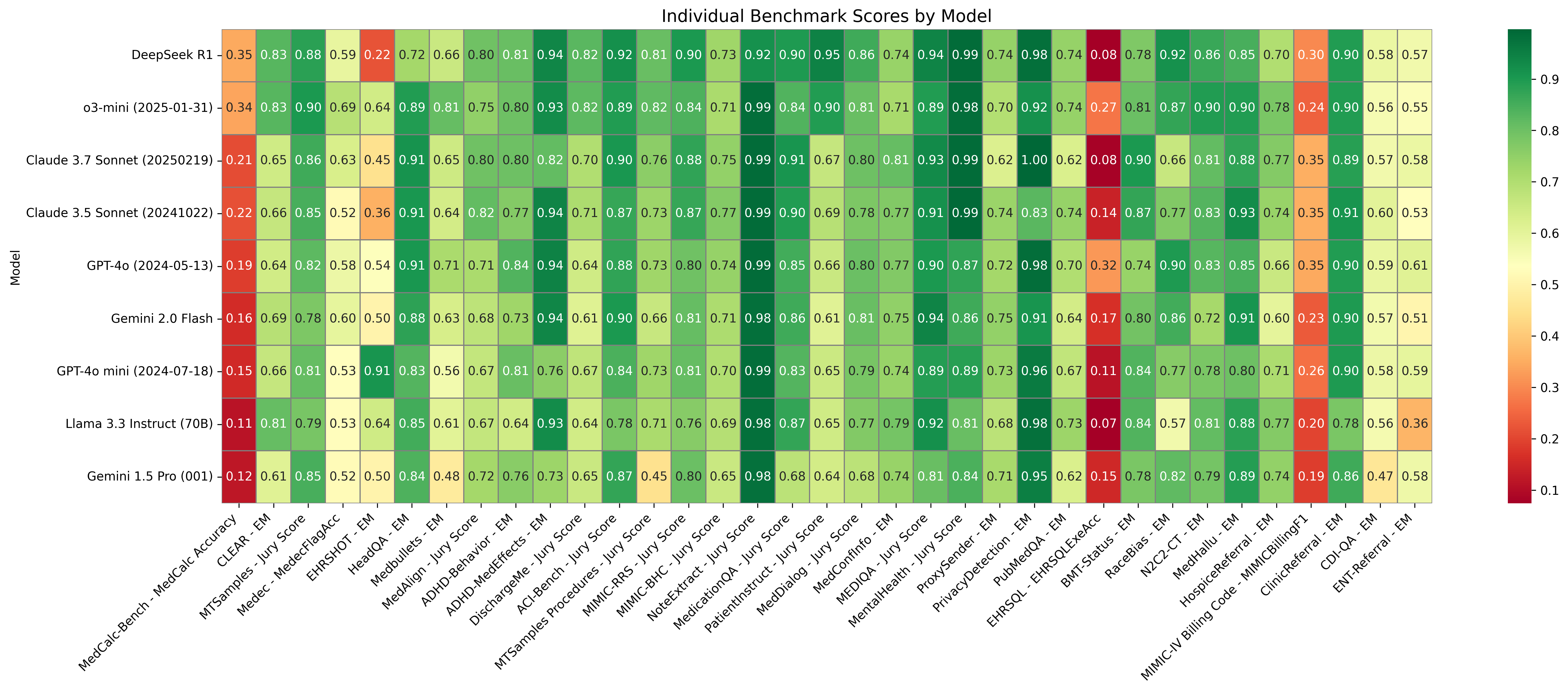}
    \caption{Heatmap of normalized scores (0–1) for each model (rows) across 35 benchmarks (columns). Scores are normalized for visualization purposes; the official leaderboard reports the original (unnormalized) scores. Dark green indicates high performance; dark red indicates low performance. The statistical significance of model differences varies by benchmark, with an analysis of minimum detectable effects shown in Appendix \ref{app:MDE}. Metrics include EM as exact match, Jury Score as the average normalized score from three frontier LLMs, MedCalc Accuracy as exact match or thresholded match depending on question type, MedFlagAcc as binary accuracy for detecting presence of errors, EHRSQLExeAcc as execution accuracy of generated code against a target output, and MIMICBillingF1 as the F1 score on ICD-10 codes extracted from a medical note.}
    \label{fig:heatmap}
\end{figure}

\paragraph{Performance by Benchmark}
We present every model's normalized score from each of the 35 benchmarks as a heatmap in Figure~\ref{fig:heatmap}, where darker green indicates higher performance.  Models perform worse on benchmarks such as MedCalc-Bench (calculating medical values from patient notes), EHRSQL (generating SQL queries from natural language instructions for clinical research, originally intended as a code generation dataset), and MIMIC-IV Billing Code (assigning ICD-10 codes to clinical notes). The best performance is seen for the NoteExtract benchmark (extracting specific information from clinical notes). Broader category-level trends are described below.

\paragraph{Performance by Category}  

\begin{figure}[t]
\centering
  \includegraphics[width=0.9\textwidth]{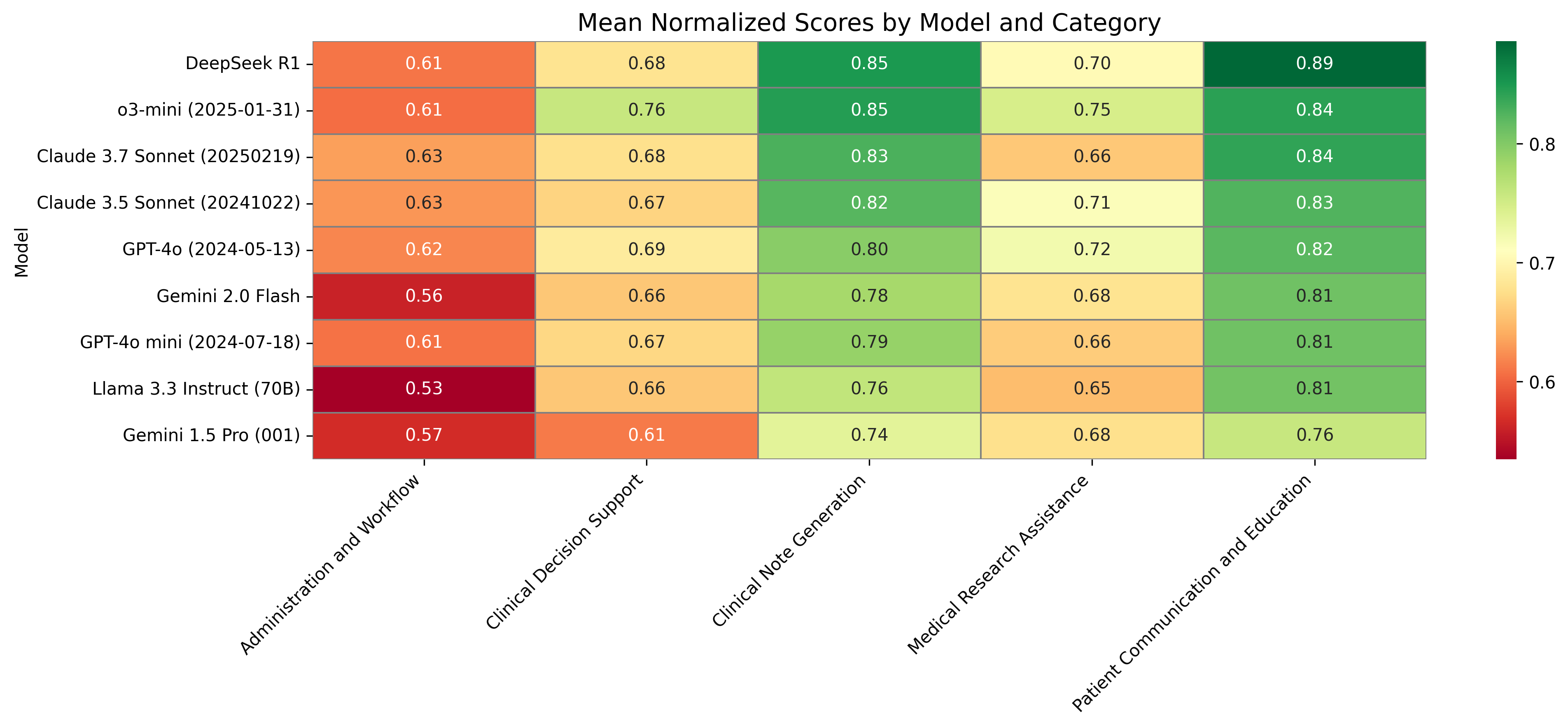}
    \caption{Mean normalized scores (0-1 scale) across the five categories for all evaluated models. Darker green represents higher scores. Models are ordered by mean win rate from top (highest) to bottom (lowest), while categories are arranged left to right.}
    \label{fig:performance-category}
    \vskip -0.8em
\end{figure}

Figure~\ref{fig:performance-category} presents performance scores by the five top-level categories in the taxonomy. Most models achieve their highest mean scores in \emph{Clinical Note Generation} \clinicalnotegenerationrange\ and \emph{Patient Communication \& Education} \patientcommunicationandeducationrange. They exhibit moderate performance in \emph{Medical Research Assistance} \medicalresearchassistancerange\ and \emph{Clinical Decision Support} \clinicaldecisionsupportrange\ and obtain generally lower scores in \emph{Administration \& Workflow} \administrationandworkflowrange. Performance variations across categories
reflect distinct task-based challenges: free-text generation tasks
(Clinical Note Generation, Patient Communication) leverage models'
natural language strengths, while structured reasoning tasks (Clinical
Decision Support, Administration) require domain-specific knowledge
integration and logical inference. These patterns have important
implications for selective deployment strategies in healthcare
settings.

At the model level, DeepSeek R1 and o3-mini lead across most categories, with DeepSeek R1 excelling in \emph{Clinical Note Generation} (0.85, tied with o3-mini) and \emph{Patient Communication \& Education} (0.89), while o3-mini leads in \emph{Clinical Decision Support} (0.76) and \emph{Medical Research Assistance} (0.75).  
The Claude Sonnet series (3.5 and 3.7) demonstrates consistent performance with scores of 0.82-0.83 in \emph{Clinical Note Generation} and 0.83-0.84 in \emph{Patient Communication \& Education}. The GPT series (4o and 4o-mini) shows moderate consistency with scores of 0.79-0.80 in \emph{Clinical Note Generation} and 0.81-0.82 in \emph{Patient Communication \& Education}.
The Gemini models exhibit greater variation, with scores of 0.74-0.78 in \emph{Clinical Note Generation} and 0.76-0.81 in \emph{Patient Communication \& Education}, where Gemini 2.0 Flash substantially outperforms 1.5 Pro. The open-source Llama 3.3 performs well in \emph{Patient Communication \& Education} (0.81) but shows the lowest score in \emph{Administration \& Workflow} (0.53), indicating areas for future improvement.

\subsubsection{Evaluation of Open-Ended Benchmarks} 
For our 13 open-ended benchmarks, we implemented an LLM-jury evaluation approach. To assess this method's validity, we collected independent clinician ratings on a subset of model outputs. We used 31 instances from ACI-Bench and 25 from MEDIQA-QA to compare clinician-assigned scores to the jury’s aggregated ratings as described in the methods. Table~\ref{tab:icc_agreement} reports intraclass correlation coefficient (ICC(3,k)) values (after rater-wise $z$-scoring) for the LLM-jury versus clinicians, alongside ROUGE-L (text overlap metric) and BERTScore-F (semantic similarity metric) baselines, and the average clinician–clinician agreement.  

Overall, the LLM-jury achieves an ICC of 0.47, which beats the average clinician--clinician agreement (ICC = 0.43) and automated baselines including ROUGE-L (0.36) and BERTScore-F (0.44). These results demonstrate that our LLM-jury mirrors clinician judgment better than standard lexical metrics, establishing its validity as a stand-in for clinician raters.  

\begin{table}[htbp]
\centering
\resizebox{\textwidth}{!}{%
\begin{tabular}{l|c|c|c|c}
\hline
\textbf{Benchmark} & \textbf{LLM} & \textbf{ROUGE-L} & \textbf{BERTScore-F} & \textbf{Clinician} \\
\hline
\texttt{Combined} & 0.474 (0.100, 0.690) & 0.361 (0, 0.630) & 0.441 (0.050, 0.670) & 0.426 (0.295, 0.585) \\
\texttt{ACI-Bench}                   & 0.305 (0, 0.670) & 0.445 (0, 0.730) & 0.250 (0, 0.640) & 0.458 (0.201, 0.945) \\
\texttt{MEDIQA}                     & 0.625 (0.150, 0.830) & 0.343 (0, 0.710) & 0.668 (0.250, 0.850) & 0.520 (0.500, 0.534) \\
\hline
\end{tabular}}
\caption{\textbf{Agreement of LLM-jury and of automated metrics with clinician ratings} 
The table entries are ICC(3,$k$) coefficients after $z$-scoring within rater (ICC3k-$z$); 95\,\% confidence intervals are shown in parentheses. 
Higher ICC indicates better agreement with clinician ratings.  
The last column gives the average clinician–clinician agreement for each dataset.}
\label{tab:icc_agreement}
\end{table}

\subsubsection{Cost–Performance Analysis}
\label{ssec:q4}

We estimated the cost (USD) of evaluating each model based on publicly listed pricing as of 05/12/2025, using the total input tokens and maximum output tokens consumed during benchmark runs and LLM-jury evaluation. These costs are an upper-bound estimate, since models may generate fewer tokens than the maximum allowed output. We plotted the mean win-rate against the cost (Figure~\ref{fig:cost-vs_winrate}, with a detailed breakdown in Table~\ref{tab:llm_comparison}, and inference costs in Appendix \ref{app:inference_costs}). As expected, non-reasoning models—GPT-4o mini (\$805) and Gemini 2.0 Flash (\$815)—incurred the lowest costs and achieved win-rates of 0.39 and 0.42, respectively. Open-source Llama 3.3 Instruct (\$940) had a 0.30 win-rate, while Gemini 1.5 Pro (\$1,130) reached 0.24. Reasoning models incurred higher costs, DeepSeek R1 (\$1,806) and o3-mini (\$1,722), with win-rates of 0.66 and 0.64, respectively. Claude 3.5 Sonnet (\$1,571) and Claude 3.7 Sonnet (\$1,537) provide a good cost-performance balance, achieving $\sim0.63$ win-rate at reduced costs.

\begin{figure}[t]
\centering
  \includegraphics[width=0.8\linewidth]{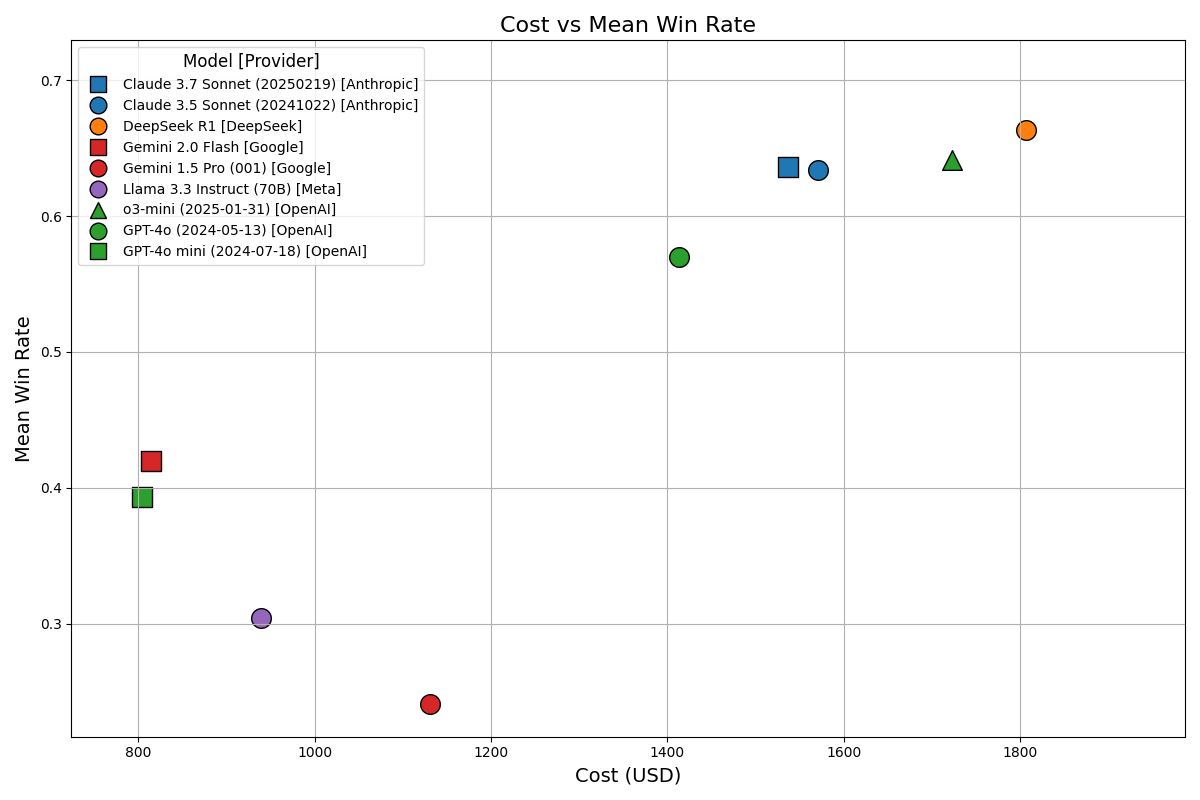}
    \caption{Scatter plot of mean win-rate (y-axis) versus estimated computational cost (x-axis) for each of the nine models across 35 benchmarks. Each point represents a model, with the position indicating the relationship between performance (y-axis) and total cost of evaluation, including benchmark runs and evaluation by LLM-jury (x-axis). Costs represent upper-bound estimates based on maximum output token usage.}
    \label{fig:cost-vs_winrate}
    \vskip -0.8em
\end{figure}

\begin{table}[ht]
  \centering
  \footnotesize
  \resizebox{\textwidth}{!}{%
    \begin{tabular}{lcccccccc}
      \toprule
      Model & Model Creator & Window Size & Access 
            & Benchmark Tokens & Jury Tokens 
            & Benchmark Cost & Jury Cost & Total Cost \\
      \midrule
      Claude 3.5 Sonnet (20241022)
        & Anthropic  & 200{,}000     & Closed
        & 245{,}958{,}343 & 45{,}868{,}483
        & \$778.67        & \$792.21   & \$1{,}570.88 \\
      Claude 3.7 Sonnet (20250219)
        & Anthropic  & 200{,}000     & Closed
        & 242{,}510{,}517 & 42{,}326{,}449
        & \$768.26        & \$768.38   & \$1{,}536.64 \\
      Gemini 1.5 Pro (001)
        & Google     & 1{,}000{,}000 & Closed
        & 277{,}273{,}571 & 42{,}864{,}352
        & \$359.33        & \$771.73   & \$1{,}131.06 \\
      Gemini 2.0 Flash
        & Google     & 1{,}000{,}000 & Closed
        & 276{,}777{,}154 & 42{,}864{,}352
        & \$43.04         & \$771.73   & \$814.77 \\
      GPT-4o (2024-05-13)
        & OpenAI     &   128{,}000   & Closed
        & 248{,}731{,}118 & 41{,}929{,}516
        & \$647.28        & \$765.91   & \$1{,}413.19 \\
      GPT-4o mini (2024-07-18)
        & OpenAI     &   128{,}000   & Closed
        & 248{,}731{,}118 & 41{,}929{,}516
        & \$38.83         & \$765.91   & \$804.74 \\
      Llama 3.3 Instruct (70B)
        & Meta AI    &   128{,}000   & Open
        & 242{,}895{,}608 & 42{,}121{,}933
        & \$172.45        & \$767.13   & \$939.58 \\
      DeepSeek R1
        & DeepSeek   &   128{,}000   & Open
        & 334{,}133{,}126 & 68{,}364{,}208
        & \$848.21        & \$957.96   & \$1{,}806.17 \\
      o3-mini (2025-01-31)
        & OpenAI     &   128{,}000   & Closed
        & 337{,}967{,}189 & 68{,}617{,}978
        & \$762.50        & \$959.54   & \$1{,}722.04 \\
      \bottomrule
    \end{tabular}%
  }
  \caption{%
    Comparison of LLMs by architecture, access type, and token usage metrics across MedHELM. Benchmark tokens denote total input/output tokens in completing the medical task represented by the benchmark; jury tokens are tokens used for open‑ended evaluations via the LLM-jury. The total cost reflects the estimated per-model expense of running a MedHELM evaluation across 35 benchmarks and represents an upper bound based on maximum output token usage. For nine models and 35 benchmarks in the current MedHELM suite, one update to the leaderboard is estimated to cost $11,739.07$.
  }
  \label{tab:llm_comparison}
  \vskip -0.5em
\end{table}

\newpage

\section{Discussion}\label{sec3}
We present a framework for assessing LLM performance for real-world medical tasks. Our clinician-validated taxonomy provides a structure for summarizing models' strengths and limitations across medical tasks. High clinician agreement (96.7\%) in assigning subcategories to categories suggests the taxonomy effectively captures how healthcare professionals conceptualize their work.
Our benchmark suite reveals nuances in model capabilities that are not seen with current medical knowledge benchmarks alone. The higher performance in communication tasks than administrative ones may stem from administrative workflows using data not seen during training, warranting caution in healthcare AI implementation for back-office tasks without quantifying task-specific performance. This framework addresses the primary limitations of current benchmarks, such as the lack of using real-world data, use of evaluation setups that do not match real-world settings, and limited task diversity.
The cost-performance trade-off shows that while reasoning models have superior performance, their substantially higher costs may not justify their deployment for all tasks. In a resource-constrained setting, models such as Claude 3.5 Sonnet offer a balance, achieving a win-rate of 0.63 at a lower cost. Finally, our LLM-jury based approach addresses a critical gap in current evaluation approaches. By beating clinician-clinician agreement, this approach enables scalable evaluation of open-ended model outputs without requiring extensive clinician time, a scarce and expensive resource.

Several limitations remain. While our LLM-jury approach was validated on only two benchmarks, expanding clinician annotations across more benchmarks would strengthen the clinician–LLM agreement estimates. In addition, the uneven distribution of benchmarks across subcategories (15 of 22 contain only one benchmark) limits our ability to draw robust performance conclusions in underrepresented areas. Moreover, our current rubrics operate at the benchmark level, but instance-level rubrics could provide better evaluation, particularly for subjective or context-dependent medical tasks where gold standard responses may not exist. Such approaches would further scale LLM-jury evaluation beyond reliance on gold standard responses \citep{croxford2025llmjudge}. Administration \& Workflow emerged as the weakest performance area for all models. Understanding the underlying causes of this poor performance, whether stemming from training data limitations, task complexity, or distributional shifts, is essential for safe deployment in healthcare operations.

In conclusion, \framework\ provides a comprehensive framework for assessing LLM performance across real-world medical tasks through our clinician-validated taxonomy spanning five categories, 22 subcategories, and 121 tasks. Our benchmark suite of 35 datasets reveals that most models perform best in Clinical Note Generation \clinicalnotegenerationrange\ and Patient Communication \& Education \patientcommunicationandeducationrange, moderately in Medical Research Assistance \medicalresearchassistancerange\ and Clinical Decision Support \clinicaldecisionsupportrange, and worst in Administration \& Workflow \administrationandworkflowrange. Reasoning models DeepSeek R1 and o3-mini led overall with win-rates of 0.66 and 0.64, respectively, though Claude models offer competitive performance (win-rate of $\sim0.63$) at lower computational cost. Through our public leaderboard\footnote{\url{https://crfm.stanford.edu/helm/medhelm/latest/}} and shared codebase\footnote{\url{https://github.com/stanford-crfm/helm}}, \framework\ establishes infrastructure for ongoing collaborative assessment as models evolve, advancing medical AI evaluation that better reflects the complexity of real-world medical practice.


\section{Methods}
\label{sec4}

\noindent
\textbf{Motivation and related work.}  
Most LLM evaluations in medicine still rely on closed‑form question‑answering over exam‑style datasets such as \textsc{MedQA} and \textsc{MedMCQA}, with only $\sim$5\% incorporating real EHR data and very few addressing free‑text generation tasks or cost‑aware metrics~\citep{medqa,medmcqa,wornow2025contextcluesevaluatinglong}.  
Large‑scale Natural Language Processing (NLP) meta‑benchmarks (\textsc{HELM}, \textsc{BIG‑Bench}) demonstrate the value of task diversity and multi‑metric scoring~\citep{liang2023holistic, bigbench}, while biomedical efforts like ClinicBench~\citep{liu2024large} or MMedBench~\citep{qiu2024towards}) each advance a single dimension (e.g.\ multimodality or cost-aware metrics) without clinician‑validated scope or extensible tooling.  Recent work on \textsc{HealthBench}~\citep{arora2025healthbench} addresses some of these limitations by incorporating physician-developed rubrics for 5,000 health conversations, but lacks a comprehensive taxonomy of medical tasks and is done with synthetic data. 
Our goal is to close these gaps by co‑designing, with clinicians, a taxonomy‑guided benchmark suite that (i) spans the full spectrum of real medical work, (ii) uses both public and private medical record data, and (iii) uses evaluation protocols that align with clinician judgment.

\subsection{Development of the Taxonomy of Tasks}
\label{subsec:taxonomy_dev}

Early attempts (\textsc{BigBIO}, ClinicBench) at creating a taxonomy of tasks in medical settings either harmonize datasets into broad categories without clinician input or collapse heterogeneous skills under a single ``generation'' label~\citep{bigbio,liu2024large}. To ensure our benchmarks accurately reflect the complexity of medical practice, we developed a taxonomy that mirrors how clinicians conceptualize their daily work. By defining a hierarchical structure, we guarantee that each benchmark maps to a concrete medical activity.
Our taxonomy consists of three levels:

\begin{itemize}
    \item \textbf{Category:} A broad domain of medical activity (e.g., \textit{Clinical Decision Support}).
    \item \textbf{Subcategory:} A group of related tasks within a Category (e.g., \textit{Supporting Diagnostic Decisions}).
    \item \textbf{Task:} A discrete action taken during the delivery of medical care  (e.g., \textit{Generate differential diagnoses}).
\end{itemize}

This structure enables systematic coverage of the medical care landscape while maintaining clear boundaries between distinct activities that care providers perform.

\subsubsection{Initial drafting}
We based our taxonomy on tasks identified in a JAMA review \cite{bedi2025testing}. Working with a clinician (MK), we reorganized these tasks into functional themes that reflect real-world activities, resulting in 98 distinct tasks organized into 21 subcategories within five categories:
\begin{enumerate}
    \item Clinical Decision Support
    \item Clinical Documentation
    \item Patient Communication \& Education
    \item Medical Research Assistance
    \item Administration \& Workflow
\end{enumerate}

Two principles guided our taxonomy development:
\begin{itemize}
    \item \textbf{Medical relevance:} Each task maps directly to actions routinely performed by care providers.
    \item \textbf{Clear boundaries:} Categories and subcategories were defined to minimize overlap while preserving meaningful functional distinctions.
\end{itemize}

\subsubsection{Validation}
To validate our initial taxonomy, we designed a two-part survey completed by 29 practicing clinicians across 14 medical specialties. More information on participating clinicians can be found in Appendix \ref{app:demographics}. The survey assessed both categorical organization and real-world relevance:

In the first section, clinicians assigned each of our 21 subcategories to one of the 5 main categories. This exercise tested whether our taxonomy structure matched how clinicians naturally organize medical tasks.

In the second section, clinicians evaluated the comprehensiveness of our taxonomy on a 5-point scale, where a score of 5 means that our categorization covered all routine medical tasks and a score of 1 means it covered very few. They also provided feedback through an open dialogue box where they could suggest missing tasks and recommend terminology improvements.
This systematic validation approach evaluated both the taxonomy's organizational logic and its comprehensiveness in representing actual medical tasks.

Based on the comments, we refined definitions and expanded the taxonomy to have 5 categories, 22 subcategories, and 121 tasks.

\subsection{Construction of the Benchmark Suite}
\label{subsec:benchmark_construction}

\subsubsection{Curation of datasets}
To construct a comprehensive suite of 35 benchmarks spanning our taxonomy, we employed a three-tiered dataset curation strategy:

\begin{enumerate}[leftmargin=1.5em]
    \item \textbf{Existing benchmarks}: We incorporated existing benchmarks from public or gated sources (e.g., MedQA, MIMIC-IV Billing Code, ACI-Bench) to ensure broad subcategory coverage.
    \item \textbf{Reformulated benchmarks}: We transformed previously unevaluated medical data collections into "reformulated benchmarks" by applying standardized prompt templates and specifying evaluation metrics. This approach addressed subcategories where datasets existed but lacked LLM-ready evaluation benchmarks.
    \item \textbf{New benchmarks}: To address the under‑representation of \emph{Administration \& Workflow}, we partnered with Stanford Healthcare to curate private datasets for tasks that are routinely done in health systems but for which benchmark datasets do not exist (e.g.\, referral triage, scheduling).  
\end{enumerate}

Each benchmark is labeled by \emph{source type} (existing / reformulated / new) and \emph{access level} (public / gated / private), with the provenance documented in the \framework\ repository.

\subsubsection{Specification of Prompts and Metrics}
To transform each curated dataset into a \framework\ benchmark, we defined three components for every item in the dataset:

\begin{itemize}
  \item \textbf{Context:} the raw input presented to the LLM (e.g.\, a clinical note, patient message, or structured EHR record).  
  \item \textbf{Prompt:} a standardized instruction template to elicit consistent, task-appropriate responses (e.g.\, “Answer in 2–3 sentences” for open-ended summaries, or MCQ framing for multiple-choice questions).  
  \item \textbf{Evaluation Metric:} a pre-specified scoring method matched to the task type:
    \begin{itemize}[nosep,leftmargin=1em]
      \item \emph{Exact-match accuracy} for single‐token or numeric outputs (e.g.\, selecting the correct option in MedQA).  
      \item \emph{Micro-F1} for multi-label classification tasks (e.g.\, ICD-10 code assignment).  
      \item \emph{LLM-jury ensemble} for open-ended text generation: we use a three-model Likert-scale protocol assessing medical accuracy, completeness and clarity, and secondary metrics, ROUGE and BERTScore, to capture lexical and semantic overlap.
    \end{itemize}
  \item \textbf{Gold Standard Response (Optional):} the reference output (numeric result, classification label, or sample text) against which the model's response is scored (e.g., "4" in response to "What's a patient's HAS-BLED score?"). While most benchmarks include a gold standard, our framework accommodates benchmarks without one, such as NoteExtract, providing flexibility for future evaluation needs.

\end{itemize}

Table~\ref{tab:headqa_case_study} illustrates these components for the HeadQA benchmark (subcategory: Providing Clinical Knowledge Support) and the MedDialog benchmark (subcategory: Patient-Provider Messaging).

We found quality issues in some of the gold standard responses for reformulated benchmarks. For example, gold standard responses in the MIMIC-RRS benchmark occasionally contained information from a patient's EHR that was not passed into the model's context.  To assess the impact of these low-quality gold standard responses, we conducted a sensitivity analysis by filtering "problematic" gold standard responses using an LLM judge and recalculating metrics (Appendix \ref{app:robustness-ablation}). Model rankings remained unchanged, as instances with "problematic" and "non-problematic" gold standard responses received similar jury scores. This stability exists because our LLM-jury (prompt in Appendix \ref{app:jury_prompt}) uses gold standard responses only when needed.

\begin{table}[ht]
\centering
\footnotesize
\renewcommand{\arraystretch}{1.2}
\setlength{\tabcolsep}{2pt}
\begin{tabular}{|l|p{4.5cm}|p{4.3cm}|p{2.4cm}|}
\hline
\rowcolor{gray!10}\textbf{Benchmark} & \textbf{Context} & \textbf{Prompt} & \textbf{\makecell[l]{Evaluation\\Metric}} \\
\hline
\textbf{HeadQA} & 
\textit{Question:} In the stools of a 6-year-old boy with diarrhea and abdominal pain, a packet of parasite eggs surrounded by a transparent membrane is observed. Which treatment do you consider most appropriate?

\smallskip
\begin{tabular}{@{}ll@{}}
\textbf{A.} Praziquantel & \textbf{B.} Azithromycin \\
\textbf{C.} Metronidazole & \textbf{D.} Albendazole
\end{tabular}
& 
You are a highly knowledgeable AI assistant specializing in biomedical sciences. Your task is to answer multiple-choice questions accurately based on the options provided. Select the correct answer by outputting only the letter corresponding to your choice (A, B, C, or D).

\smallskip
\textbf{Question:} \{Context\}

\textbf{Answer:}
& 
Exact Match

\smallskip
\textbf{Answer: D}
\\
\hline
\textbf{MedDialog} & 
\textit{Patient:} I have had this pain in my left arm down to almost my wrist. It first started where I would have a sharp pain between my armpit and boob. It doesn't hurt all the time but has become more often lately. It’s an aching throbbing feeling but also a numbing weak feeling to where I lose my strength in my arm. After a while it will pass but I don't know if it's something serious. Need help.

\smallskip
\textit{Doctor:} Hi, take this seriously, it could be a lump compressing on your nerves, or any infection. You need an examination by a doctor to exactly judge what your symptoms mean and what is their cause. I recommend you consult your doctor at the earliest.
& 
Generate a one sentence summary of this patient-doctor conversation.
& 
LLM Jury

\smallskip
\textbf{Accuracy: 4.88}
\\
\hline
\end{tabular}
\caption{Case study of HeadQA and MedDialog, two benchmarks under MedHELM}
\label{tab:headqa_case_study}
\end{table}

\subsection{Model Evaluation and Cost-Performance Analysis}
\label{subsec:model_eval_methods}

\subsubsection{Model Selection and Inference Pipeline}
We evaluated 9 state-of-the-art LLMs (Table~\ref{tab:model_macroavg}) under a uniform prompting and decoding regimen. All models were queried via their respective APIs or local endpoints with sampling temperature set to 0 for deterministic outputs. All experiments were conducted on a PHI-compliant shared cluster maintaining full HIPAA compliance.

\subsubsection{Performance Metrics}
To quantify task performance, we computed:
\begin{itemize}[nosep]
  \item \textbf{Pairwise win-rate:} For each of the 35 benchmarks, we compared each model against every other; a ``win'' is assigned if a model's normalized score $\geq$ its rival's. We then averaged wins over all pairings.
  \item \textbf{Macro-average score:} The overall performance score calculated by averaging results across all 35 benchmarks, with each benchmark weighted equally regardless of size (0–1 scale).
\end{itemize}

\subsubsection{Evaluation of Open-Ended Benchmarks}
\label{subsubsec:open_ended_eval}

For open-ended benchmarks, early pipelines used a single “LLM‑as‑Judge” to score outputs \citep{gu2025surveyllmasajudge,Samuylova2025LLMJudge}, but high variance and bias spurred the “LLM‑as‑Jury” paradigm, aggregating $k$ independent judgments for closer expert agreement \citep{madaan2024quantifyingvarianceevaluationbenchmarks,verga2024replacingjudgesjuriesevaluating}. Extensions like G‑Eval introduce chain-of-thought (CoT) self‑critique per juror, and tools such as SelfCheckGPT and FActScore target hallucination and factuality \citep{deepeval2024,manakul2023selfcheckgptzeroresourceblackboxhallucination,min2023factscorefinegrainedatomicevaluation}. \framework\ adopts a three‑member jury without CoT to balance reliability and runtime.

\paragraph{Evaluation using LLM-Jury}  
We selected a three-member jury based on
prior research demonstrating that odd-numbered panels reduce tie
scenarios while maintaining reliability \cite{guha2016jury}. The jury composition (GPT-4o, Claude 3.7 Sonnet, LLaMA 3.3 70B) was chosen to represent diverse
model architectures and training approaches, minimizing systematic
bias from any single provider. We prompted each judge to score the model-generated responses on a 1–5 Likert scale according to three axes adopted from \cite{vanveen2024adapted}:
\begin{itemize}[nosep]
  \item \textbf{Accuracy:} Factual correctness and adherence to medical guidelines.  
  \item \textbf{Completeness:} Thoroughness in addressing all aspects of the query.  
  \item \textbf{Clarity:} Organization, readability, and easy to understand language.  
\end{itemize}
For the NoteExtract benchmark, which requires restructuring free-form clinical care plans into a specified format without a gold standard response, we modified our evaluation approach. We replaced the "completeness" criterion with "structure," which assessed whether model responses properly reorganized the input text according to the requested format.
The final LLM-jury score for each response is the mean of all nine ratings (3 judges × 3 axes).

\paragraph{Clinician Rating}  
To validate the LLM-jury approach, we collected clinician ratings on a subset of two open-ended benchmarks (MEDIQA and ACI-Bench).  We selected these benchmarks because they were publicly available (facilitating annotation) and represent different categories (Patient Communication and Education and Clinical Note Generation, respectively). 20 clinicians across various specialties each scored a subset of responses on the same three axes, with at least two clinicians per instance. 

\paragraph{Clinician–LLM Agreement Metrics}  
We assessed agreement between the LLM-jury and clinicians via two intraclass correlation coefficients (ICC), after applying z-score normalization to each rater’s composite (mean of the three axis ratings) to remove scale biases:
\[
z_{ij} = \frac{x_{ij} - \bar{x}_i}{s_i}\,,
\]
where \(x_{ij}\) is rater \(i\)’s composite on instance \(j\).  

\begin{enumerate}
    \item \textbf{ICC(3,k)\(_\text{z}\)}: We treat the clinicians' mean and LLM jury mean as two fixed raters in a two-way mixed-effects model, thus, we compute  
\[
\mathrm{ICC}(3,k)
  = \frac{MS_\text{cases} - MS_\varepsilon}{MS_\text{cases}}, \quad k=2,
\]
to quantify absolute agreement.  
    \item \textbf{Average Clinician–Clinician ICC\(_\text{z}\)}: For every pair of clinicians who scored at least two common instances, we z-normalize each rater’s composite scores and compute ICC(3,k=2). We then average these pairwise ICCs (and bootstrap a 95 \% CI) to give a representative inter-clinician agreement baseline.

Together, these metrics measure (i) the fidelity of LLM-jury to expert clinician rating and (ii) whether LLM–Clinician alignment approaches inter-clinician agreement. 
\end{enumerate}

ROUGE and BERTScore were also computed for all open-ended benchmarks with gold
standard responses but used only as secondary metrics due to their limited alignment with
clinical judgment.

\subsubsection{Cost–Performance Analysis}
We tracked token usage for both benchmark prompts and jury evaluations across all models by counting input tokens and assuming the maximum allowed output tokens. This approach was necessary because, for o3-mini, it is not possible to fully observe the token usage as the thinking tokens don't appear in the final output. As a result, our cost estimates represent an upper bound. Using published per-1M-token pricing as of 05/12/2025, we estimated the total cost per model. Cost–performance trade-offs were visualized by plotting mean win-rate against estimated total evaluation cost, highlighting models that optimize accuracy per dollar spent.

\medskip
\noindent
By integrating clinician‑validated taxonomy design, a balanced mix of public and private datasets, and LLM‑jury evaluation, our benchmark addresses the three structural gaps identified in earlier work—task diversity, medical grounding, and metric fidelity—while providing transparent cost accounting for real‑world deployment decisions.

\newpage

\begin{appendices}

\appendix
\section{Task Taxonomy}
\label{app:taxonomy}

\subsection*{Table of Contents}
\begin{enumerate}
    \item Overview \dotfill 1
    \item Clinical Decision Support \dotfill 2  
    \begin{itemize}
        \item Definition
        \item Subcategories and Tasks
    \end{itemize}
    \item Clinical Note Generation \dotfill 3
    \begin{itemize}
        \item Definition
        \item Subcategories and Tasks
    \end{itemize}
    \item Patient Communication and Education \dotfill 5
    \begin{itemize}
        \item Definition
        \item Subcategories and Tasks
    \end{itemize}
    \item Medical Research Assistance \dotfill 7
    \begin{itemize}
        \item Definition
        \item Subcategories and Tasks
    \end{itemize}
    \item Administration and Workflow \dotfill 9
    \begin{itemize}
        \item Definition
        \item Subcategories and Tasks
    \end{itemize}
\end{enumerate}

\subsection*{Overview}
\begin{itemize}
    \item \textbf{Total Categories:} 5
    \item \textbf{Entities per Category:}
    \begin{itemize}
        \item \textbf{Definition:} A description of the category.
        \item \textbf{Inclusion criteria:} Guidelines determining which subcategories (and tasks) belong in the category.
        \item \textbf{Task performer:} Individuals responsible for executing tasks under the category.
        \item \textbf{Subcategories:} Each category contains multiple subcategories.
        \item \textbf{Tasks:} Each subcategory comprises several tasks.
    \end{itemize}
\end{itemize}


        \begin{verbatim}
        Category
        - Definition
        - Inclusion criteria
        - Task performer
        - Sub category
            - Task
            - Task
            - Task
        - Sub category
            - Task
            - Task
            - Task
        \end{verbatim}

\subsection*{1. Clinical Decision Support}

\textbf{Definition:} Analyzing patient-specific data to provide evidence-based recommendations to clinicians.

\noindent\textbf{Inclusion criteria:} Actions generating actionable insights based on patient data and medical evidence.

\noindent\textbf{Task performer:} Healthcare professionals (clinicians, nurses, and other practitioners).  
\newline

\noindent\textbf{Subcategories and Tasks:}
\begin{itemize}
    \item \textbf{Supporting Diagnostic Decisions}
    \begin{itemize}
        \item Recognize disease patterns from symptoms/vitals
        \item Interpret diagnostic tests (ECG, spirometry)
        \item Generate follow-up questions
        \item Generate differential diagnoses
        \item Interpret lab results
        \item Detect image findings
        \item Perform medical calculations
        \item Evaluate social determinants of health
        \item Track lab trends
        \item Process intake information
    \end{itemize}
    \item \textbf{Planning Treatments}
    \begin{itemize}
        \item Check drug interactions
        \item Match protocols / screen contraindications
        \item Suggest clinical pathways
        \item Predict treatment response
        \item Make collaborative decisions
        \item Evaluate treatment accessibility
    \end{itemize}
    \item \textbf{Predicting Risks and Outcomes}
    \begin{itemize}
        \item Predict deterioration, readmission, disease progression
        \item Predict outcomes, adverse events, discharge readiness
        \item Predict need for procedures or referrals
        \item Manage preventive screening
    \end{itemize}
    \item \textbf{Providing Clinical Knowledge Support}
    \begin{itemize}
        \item Apply guidelines and best practices
        \item Answer medical knowledge questions
        \item Track protocol compliance
        \item Assess care quality
    \end{itemize}
\end{itemize}

\subsection*{2. Clinical Note Generation}
\textbf{Definition:} Creating structured records of patient care. 

\noindent\textbf{Inclusion criteria:} Actions producing or modifying official clinical records.  

\noindent\textbf{Task performer:} Providers, scribes, and documentation specialists.  
\newline

\noindent\textbf{Subcategories and Tasks:}
\begin{itemize}
    \item \textbf{Documenting Patient Visits}
    \begin{itemize}
        \item Generate progress, consultation, ED, admission, and discharge notes
        \item Synthesize external/internal records
        \item Summarize clinical documents
        \item Generate team assessments
    \end{itemize}
    \item \textbf{Recording Procedures}
    \begin{itemize}
        \item Generate OR, bedside, specialized procedure notes
    \end{itemize}
    \item \textbf{Documenting Diagnostic Reports}
    \begin{itemize}
        \item Generate imaging, pathology, test, and genomic reports
    \end{itemize}
    \item \textbf{Documenting Care Plans}
    \begin{itemize}
        \item Document treatment plans, care protocols, nursing plans, advance planning
    \end{itemize}
\end{itemize}

\subsection*{3. Patient Communication and Education}
\textbf{Definition:} Transmitting health information to enable patient understanding.  

\noindent\textbf{Inclusion criteria:} Acts that support informed patient participation.  

\noindent\textbf{Task performer:} Providers, coordinators, educators.  
\newline

\noindent\textbf{Subcategories and Tasks:}
\begin{itemize}
    \item \textbf{Providing Education Resources}
    \begin{itemize}
        \item Simplify disease info, risk factors, treatment
        \item Explain insurance and billing
    \end{itemize}
    \item \textbf{Delivering Personalized Instructions}
    \begin{itemize}
        \item Generate medication, procedure, and home care instructions
        \item Explain follow-up and recovery
    \end{itemize}
    \item \textbf{Patient-Provider Messaging}
    \begin{itemize}
        \item Triage messages, analyze symptoms
        \item Handle refills, appointments, questions
        \item Share results, draft responses
    \end{itemize}
    \item \textbf{Enhancing Accessibility}
    \begin{itemize}
        \item Generate visual aids, translate content, make content accessible
    \end{itemize}
    \item \textbf{Facilitating Engagement}
    \begin{itemize}
        \item Send reminders, preventive care, track goals
        \item Collect feedback, support counseling and groups
    \end{itemize}
\end{itemize}

\subsection*{4. Medical Research Assistance}
\textbf{Definition:} Analyzing clinical data and literature to advance medical knowledge.  

\noindent\textbf{Inclusion criteria:} Actions transforming data into scientific evidence.  

\noindent\textbf{Task performer:} Researchers and epidemiologists.  
\newline

\noindent\textbf{Subcategories and Tasks:}
\begin{itemize}
    \item \textbf{Conducting Literature Research}
    \begin{itemize}
        \item Screen reviews, summarize papers
        \item Analyze citations, synthesize evidence, identify gaps
    \end{itemize}
    \item \textbf{Analyzing Research Data}
    \begin{itemize}
        \item Analyze trials, compare treatments, assess outcomes
        \item Conduct cohort studies
    \end{itemize}
    \item \textbf{Recording Research Processes}
    \begin{itemize}
        \item Support protocols, grants, manuscripts, statistics
    \end{itemize}
    \item \textbf{Ensuring Research Quality}
    \begin{itemize}
        \item Validate methods, assess bias, handle regulatory needs
    \end{itemize}
    \item \textbf{Managing Enrollment}
    \begin{itemize}
        \item Screen and match patients, track and document recruitment
    \end{itemize}
\end{itemize}

\subsection*{5. Administration and Workflow}
\textbf{Definition:} Orchestrating clinical operations from scheduling to billing.  

\noindent\textbf{Inclusion criteria:} Actions managing logistics, resources, and flow.  

\noindent\textbf{Task performer:} Administrators, staff, billing specialists.  
\newline

\noindent\textbf{Subcategories and Tasks:}
\begin{itemize}
    \item \textbf{Scheduling Resources and Staff}
    \begin{itemize}
        \item Schedule staff, manage inventory, equipment, and facilities
        \item Monitor institutional metrics
    \end{itemize}
    \item \textbf{Overseeing Financial Activities}
    \begin{itemize}
        \item Generate/document billing, insurer communication
        \item Analyze revenue/costs, estimate out-of-pocket costs
    \end{itemize}
    \item \textbf{Organizing Workflow Processes}
    \begin{itemize}
        \item Schedule appointments, process referrals/documents
        \item Handle information requests
    \end{itemize}
    \item \textbf{Care Coordination and Planning}
    \begin{itemize}
        \item Evaluate admissions, coordinate providers
        \item Manage post-discharge planning and transitions
    \end{itemize}
\end{itemize}

\section{Clinician Participant Demographics}
\label{app:demographics}

We validated our task taxonomy through a survey of 29 clinicians representing 14 medical specialties across 4 institutions. Tables \ref{tab:speciality_counts} and \ref{tab:affiliation_counts} provide detailed breakdowns of participant specialties and affiliations.

\begin{table}[htbp]
\centering
\begin{tabular}{ll}
\toprule
\textbf{Affiliation} & \textbf{Number of Participants} \\
\midrule
Stanford University    & 26 \\
Mayo Clinic     & 1 \\
Oregon Health \& Science University     & 1 \\
Flourish Research     & 1 \\
\bottomrule
\end{tabular}
\caption{Number of participants per affiliation.}
\label{tab:affiliation_counts}
\end{table}

\begin{table}[htbp]
\centering
\begin{tabular}{ll}
\toprule
\textbf{Speciality} & \textbf{Number of Participants} \\
\midrule
Internal Medicine     & 9 \\
Radiology             & 2 \\
Otolaryngology        & 2 \\
Neurology             & 2 \\
Family Medicine       & 2 \\
Dermatology           & 2 \\
Anesthesiology        & 2 \\
Radiation Oncology    & 2 \\
Emergency Medicine    & 1 \\
Ophthalmology         & 1 \\
Psychiatry            & 1 \\
Pathology             & 1 \\
Pediatrics            & 1 \\
Nuclear Medicine      & 1 \\
\bottomrule
\end{tabular}
\caption{Number of participants per speciality.}
\label{tab:speciality_counts}
\end{table}

\section{List of Benchmarks in MedHELM}
\label{app:benchmark}
\vspace{-1.5em}
\begin{table}[!htbp]
  \centering
  \footnotesize
  \renewcommand{\arraystretch}{1.3}
  \setlength{\tabcolsep}{3pt}
  \resizebox{\linewidth}{!}{%
  \begin{tabular}{|p{2.8cm}|p{3.2cm}|p{7.7cm}|p{2cm}|p{2cm}|}
  \hline
  \rowcolor{gray!10} \textbf{Category} & \textbf{Dataset Name} & \textbf{Dataset Description} & \textbf{Access Level} & \textbf{\makecell[l]{Curation\\Status}}\\
  \hline
  
  \multirow{10}{*}{\textbf{\makecell[l]{Clinical Decision\\Support}}}
  & MedCalc-Bench &  A dataset which consists of a patient note, a question requesting to compute a specific medical value, and a ground truth answer. & Public & Existing \\
  & CLEAR & A dataset that evaluates medical condition detection from patient notes using yes/no/maybe classifications. & Private & Formulated \\
  & MTSamples & A dataset that provides transcribed medical reports and prompts models to generate appropriate treatment plans. & Public & Formulated \\
  & Medec &   A dataset containing medical narratives with error detection and correction pairs. & Public & Existing \\
  & EHRSHOT &  A dataset given a patient record of EHR codes, classifying if an event will occur at a future date or not. & Gated & Existing \\
  & HeadQA &  A collection of biomedical multiple-choice questions for testing medical knowledge. & Public & Existing \\
  & Medbullets &  A USMLE-style medical question dataset with multiple-choice answers and explanations. & Public & Existing \\
  & MedAlign &  A dataset that asks models to answer questions/follow instructions over longitudinal EHR. & Gated & Existing \\
  & ADHD-Behavior &  A dataset that classifies whether a clinical note contains a clinician recommendation for parent training in behavior management, which is the first-line evidence-based treatment for young children with ADHD. & Private & New \\
  & ADHD-MedEffects &  A dataset that classifies whether a clinical note contains documentation of side effect monitoring (recording of absence or presence of medication side effects), as recommended in clinical practice guidelines.  & Private & New \\

  \hline
  
  \multirow{6}{*}{\textbf{\makecell[l]{Clinical Note\\Generation}}} 
  & DischargeMe & A dataset that provides discharge text as well as the radiology report text collected from MIMIC-IV data such that models can generate discharge instructions and brief hospital course information generation. & Gated & Existing \\
  & ACI-Bench & A dataset of patient-doctor conversations paired with structured clinical notes. & Public & Existing \\
  & MTSamples Procedures & A dataset that provides a patient note regarding an operation, with the objective to document the procedure. & Public & Formulated \\
  & MIMIC-RRS & A dataset containing radiology reports with findings sections from MIMIC-III paired with their corresponding impression sections, used for generating radiology report summaries. & Gated & Formulated \\
  & MIMIC-BHC & A summarization task using a curated collection of preprocessed discharge notes paired with their corresponding brief hospital course (BHC) summaries. & Gated & Existing \\
  & NoteExtract & A dataset containing free form text of a clinical health worker care plan, with the associated goal being to restructure that text into a given format. & Private & New \\

  \hline
  
  \multirow{8}{*}{\textbf{\makecell[l]{Patient\\Communication\\and Education}}}
  & MedicationQA & A dataset containing open text question-answer pairs regarding consumer health questions about medication. & Public & Existing \\
  & PatientInstruct & A dataset containing case details used to generate customized post-procedure patient instructions. & Private & New \\
  & MedDialog & A collection of doctor-patient conversations with corresponding summaries. & Public & Existing \\
  & MedConfInfo & A dataset of clinical notes from adolescent patients used to identify sensitive protected health information that should be restricted from parental access. & Private & New \\
  & MEDIQA-QA & A dataset including a medical question, a set of candidate answers,relevance annotations for ranking, and additional context to evaluate understandingand retrieval capabilities in a medical setting. & Public & Existing \\
  & MentalHealth & A dataset containing a counselor and mental health patient conversation, where the objective is to generate an empathetic counselor response. & Private & New \\
  & PrivacyDetection & A dataset that determines if a message leaks any confidential information from the patient & Private & New \\
  & ProxySender & A dataset that determines if a message was sent by a proxy user & Private & New \\

  \hline
  
  \multirow{6}{*}{\textbf{\makecell[l]{Medical Research\\Assistance}}} 
  & PubMedQA & A dataset that provides pubmed abstracts and asks associated questions yes/no/maybe questions. & Public & Existing \\
  & EHRSQL & A dataset that generates an SQL query that would be used in clinical research given a natural language instruction. & Public & Existing \\
  & BMT-Status & A dataset containing patient notes with associated questions and answers related to bone marrow transplantation. & Private & New \\
  & RaceBias & A collection of LLM outputs in response to medical questions with race-based biases, with the objective being to classify whether the output contains racially biased content. & Public & Formulated \\
  & N2C2-CT & A dataset that provides clinical notes and asks the model to classify whether the patient is a valid candidate for a provided clinical trial. & Gated & Existing \\
  & MedHallu & A dataset of PubMed articles and associated questions, with the objective being to classify whether the answer is factual or hallucinated. & Public & Existing \\
  \hline
  
  \multirow{5}{*}{\textbf{\makecell[l]{Administration\\and Workflow}}} 
  & HospiceReferral & A dataset evaluating performance in identifying appropriate patient referrals to hospice care. & Private & New \\
  & MIMIC-IV Billing Code & A dataset pairing clinical notes from MIMIC-IV with corresponding ICD-10 billing codes. & Gated & Existing \\
  & ClinicReferral & A dataset containing manually curated answers to questions regarding patient referrals to the Sequoia clinic. & Private & New \\
  & CDI-QA & A dataset built from Clinical Document Integrity (CDI) notes, to assess the ability to answer verification questions from previous notes. & Private & New \\
  & ENT-Referral & A dataset designed to evaluate performance in identifying appropriate patient referrals to Ear, Nose, and Throat specialists. & Private & New \\
  \hline
  
  \end{tabular}
  }
  \caption{Overview of all 35 benchmarks included in MedHELM, categorized by category, access level and curation status.}
  \label{tab:new_datasets}
  \end{table}

\newpage

\section{LLM-based filtering}
\label{app:robustness-ablation}
We assessed the quality of gold standard responses for reformulated benchmarks. Using GPT-4o-mini as a classifier, we identified reasonable versus unreasonable gold standard responses with the following prompt: \fbox{\parbox{\textwidth}{%
Evaluate the following text: \{prompt\}\{context\} against the following 
reference: \{gold standard response\}. Return a score of 0 if the gold standard response appears 
to be using outside information that is not present in the input text 
and is not expected to be external knowledge known by a clinician, 
and thus is an unreasonable gold standard response to the question. 
Give a score of 1 if the reference is a reasonable response to the 
prompt. Only give a score of 0 if the reference does not make sense 
with the prompt or the question couldn't be answered in this way 
without additional information. Only return the score, no other text.
}}

\begin{figure}[!htbp]
\centering
  \includegraphics[width=0.9\textwidth]{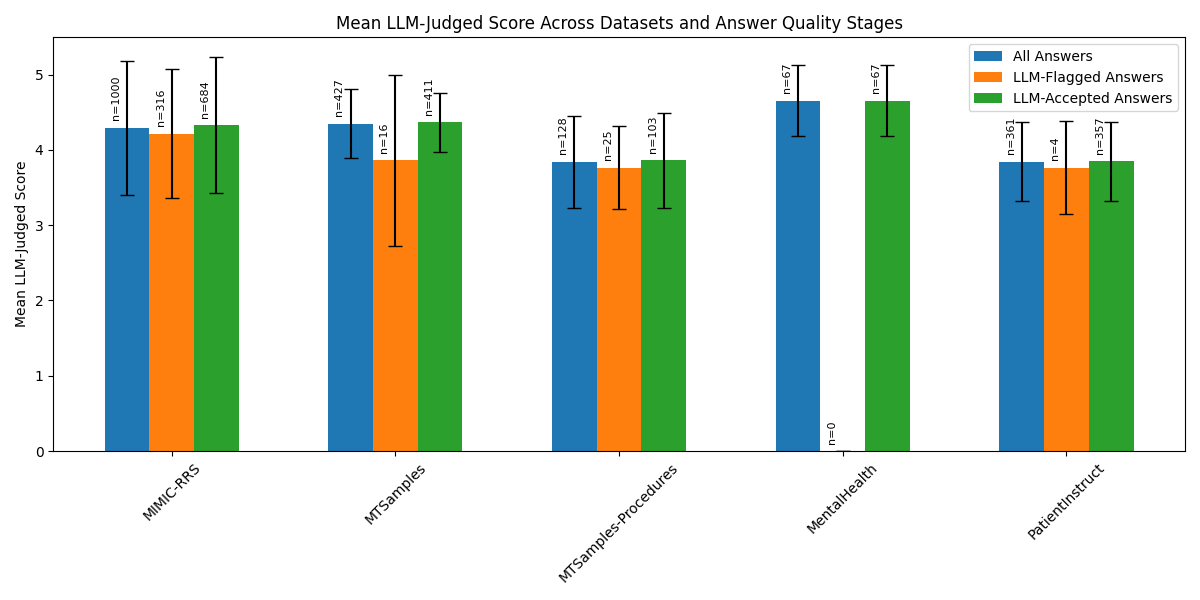}
    \caption{We filter the question-answer pairs into a positive and negative group via LLM-based filtering. We see that the mean score of both groups remains relatively consistent, indicating stability in judging.}
    \label{fig:robustness}
\end{figure}

As shown in Figure ~\ref{fig:robustness}, 3 out of 5 reformulated benchmarks have less than 5\% detected error in the gold standard response. For the remaining 2 benchmarks (MIMIC-RRS and MTSamples-Procedures), our evaluation methodology proved robust to gold standard response imperfections, with consistent mean scores between flagged and accepted answers. This resilience stems from our LLM-jury prompt design, which instructs judges to reference gold standards only when necessary.

\section{Minimum Detectable Effect Evaluations}
\label{app:MDE}
We calculate the minimum detectable effect at the benchmark level with the following formulation, where \textit{b} is the benchmark, \textit{ij} are two models to compare in a paired evaluation selected from all 9 models in $M$, $\sigma_b^{ij}$ is the standard deviation of the difference between the two selected model outputs for every question in the benchmark, and $n_b$ is the number of questions in the benchmark. $\mathrm{SD}_b$ is the standard deviation over the total pairwise MDE scores for the given benchmark, $b$. Additionally, we set $\alpha=0.05$ and $\beta=0.20$
\[
\mathrm{MDE}_b = \frac{1}{|\mathcal{M}|} \sum_{(i, j) \in \mathcal{M}} \frac{(z_{1-\frac{\alpha}{2}} + z_{1-\beta}) \cdot \sigma_{b}^{(i,j)}}{\sqrt{n_b}}
\quad \pm \quad \mathrm{SD}_b
\]

Table~\ref{tab:mde_benchmarks} shows the MDE for each benchmark and supports that the differences identified between model performance are significant.

\begin{table}[!htbp]
\centering
\begin{tabular}{l|c}
\hline
\textbf{Benchmark} & \textbf{MDE} \\
\hline
PatientInstruct & $0.008 \pm 0.002$ \\
MIMIC-RRS & $0.018 \pm 0.005$ \\
CLEAR & $0.054 \pm 0.003$ \\
ENT-Referral & $0.045 \pm 0.005$ \\
NoteExtract & $0.009 \pm 0.011$ \\
CDI-QA & $0.050 \pm 0.007$ \\
N2C2 & $0.087 \pm 0.010$ \\
MentalHealth & $0.032 \pm 0.011$ \\
PrivacyDetection & $0.040 \pm 0.013$ \\
ProxySender & $0.074 \pm 0.009$ \\
MIMIC-IV-BHC & $0.019 \pm 0.006$ \\
ADHD-Behavior & $0.056 \pm 0.008$ \\
ClinicReferral & $0.050 \pm 0.011$ \\
DischargeMe & $0.008 \pm 0.002$ \\
MedConfInfo & $0.031 \pm 0.004$ \\
EHRSHOT & $0.033 \pm 0.004$ \\
MIMIC-IV Billing Code & $0.015 \pm 0.003$ \\
BMT-Status & $0.079 \pm 0.009$ \\
HospiceReferral & $0.042 \pm 0.005$ \\
ADHD-MedEffects & $0.032 \pm 0.009$ \\
MedAlign & $0.041 \pm 0.005$ \\
EHRSQL & $0.034 \pm 0.009$ \\
MTSamples Procedures & $0.020 \pm 0.010$ \\
ACI-Bench & $0.012 \pm 0.002$ \\
MedicationQA & $0.012 \pm 0.004$ \\
MedCalc-Bench & $0.037 \pm 0.005$ \\
MEDIQA & $0.018 \pm 0.008$ \\
MedBullets & $0.079 \pm 0.006$ \\
MedDialog & $0.007 \pm 0.002$ \\
MTSamples & $0.011 \pm 0.002$ \\
Medec & $0.052 \pm 0.011$ \\
PubMedQA & $0.037 \pm 0.004$ \\
HeadQA & $0.030 \pm 0.007$ \\
MedHallu & $0.030 \pm 0.004$ \\
RaceBias & $0.092 \pm 0.016$ \\
\hline
\end{tabular}
\caption{Minimum Detectable Effect (MDE) values for various benchmarks}
\label{tab:mde_benchmarks}
\end{table}

\section{LLM-Jury Prompt}
\label{app:jury_prompt}
While each jury prompt is customized for the given benchmark, we include the ACI-Bench jury prompt to serve as an example of how our LLM-Jury prompt is structured. 
\begin{lstlisting}
You are a medical expert tasked with evaluating the quality of a
generated clinical note based on a doctor-patient conversation.
Your goal is to assess how well the note captures the clinical information from the conversation and
compare it to the reference note (gold standard) in terms of accruacy, completeness and clarity.
The conversation will be provided in these tags:
<conversation>{{QUESTION}}</conversation>
The generated note will be provided in these tags:
<response>{{RESPONSE}}</response>
The reference note will be provided in these tags:
<gold_response>{{GOLD_RESPONSE}}</gold_response>
Carefully review the <response> based on the <conversation> and compare it to the <gold_response> when needed.
For each of the following criteria, rate the response on a scale of 1 to 5 (1 = very poor, 5 = excellent),
and provide a short justification for your score.
Evaluation Criteria:
Accuracy (1-5) - Does the note provide correct clinical information based on the conversation?
Completeness (1-5) - Does the note include all important medical details from the conversation?
Clarity (1-5) - Is the note written clearly and organized in a standard clinical format for clinicians
Output Format:
Output the evaluation as a single valid JSON object matching the following structure:
{"accuracy":{"score":0,"explanation":"Explain why this score was given."},
"completeness":{"score":0,"explanation":"Explain why this score was given."},
"clarity":{"score":0,"explanation":"Explain why this score was given."}}
Ensure the output is valid JSON:
- Use **double quotes** (") for all keys and string values.
- When quoting text or sections inside the explanations, use escaped double quotes (\") to
  maintain valid JSON formatting.
- Do not include any additional information in the output.
\end{lstlisting}

\newpage
\section{LLM Inference Costs}
\label{app:inference_costs}

Table \ref{tab:inference_costs} summarizes the per-token inference costs of the LLMs used in MedHELM as of 05/12/2025. 

\begin{table}[htpb]
\centering
\begin{tabular}{lcccc}
\toprule
\multirow{2}{*}{\textbf{Model}} & \multirow{2}{*}{\textbf{Release Date}} & \multicolumn{2}{c}{\textbf{Cost (per million tokens)}} \\
\cmidrule(lr){3-4}
 & & \textbf{Input Tokens} & \textbf{Output Tokens} \\
\midrule
\href{https://openai.com/api/pricing/}{GPT-4o} & 05/13/2024 & \$2.50 & \$10.00 \\
\href{https://openai.com/api/pricing/}{GPT-4o Mini} & 07/18/2024 & \$0.15 & \$0.60 \\
\href{https://azuremarketplace.microsoft.com/en-us/marketplace/apps/metagenai.llama-3-3-70b-instruct-offer?tab=PlansAndPrice}{Llama 3.3 Instruct (70B)} & 12/06/2024 & \$0.71 & \$0.71 \\
\href{https://cloud.google.com/vertex-ai/generative-ai/pricing}{Gemini 2.0 Flash} & 02/01/2025 & \$0.15 & \$0.60 \\
\href{https://cloud.google.com/vertex-ai/generative-ai/pricing}{Gemini 1.5 Pro (001)} & 05/24/2024 & \$1.25 & \$5.00 \\
\href{https://aws.amazon.com/bedrock/pricing/}{Claude 3.5 Sonnet} & 10/22/2024 & \$3.00 & \$15.00 \\
\href{https://aws.amazon.com/bedrock/pricing/}{Claude 3.7 Sonnet} & 02/19/2025 & \$3.00 & \$15.00 \\
\href{https://azure.microsoft.com/en-us/pricing/details/cognitive-services/openai-service/}{o3-mini} & 01/31/2025 & \$1.20 & \$4.84 \\
\href{https://ai.azure.com/explore/models/DeepSeek-R1/version/1/registry/azureml-deepseek}{DeepSeek R1} & 01/20/2025 & \$1.20 & \$4.84 \\
\bottomrule
\end{tabular}
\caption{\textbf{Per-token inference costs of LLMs used in MedHELM.}}
\label{tab:inference_costs}
\end{table}

\end{appendices}



\clearpage
\bibliography{sn-bibliography}

\end{document}